\documentclass[11pt]{article}

\usepackage[preprint]{acl}

\usepackage{times}
\usepackage{latexsym}
\usepackage{enumitem}
\usepackage{amssymb}
\usepackage[T1]{fontenc}

\usepackage[utf8]{inputenc}

\usepackage{microtype}

\usepackage{inconsolata}

\usepackage{graphicx}

\usepackage{xcolor}
\usepackage{color}

\usepackage{CJKutf8}
\usepackage{graphicx} 
\usepackage{caption} 

\usepackage{lineno} 

\usepackage{colortbl}
\usepackage{xcolor}
\usepackage[most]{tcolorbox}

\usepackage{booktabs}
\definecolor{jirailight}{RGB}{227,214,231}    
\definecolor{jiraimedium}{RGB}{204,181,200}   
\definecolor{jiraipink}{RGB}{182,136,158}     
\definecolor{jiraimid}{RGB}{130,116,139}      
\definecolor{jiraidark}{RGB}{54,45,59}        

\title{Towards Valid Student Simulation with Large Language Models}

\author{
  Zhihao Yuan$^{1}$\thanks{Equal contribution.}, Yunze Xiao$^{1}$\footnotemark[1], Ming Li$^{2}$\footnotemark[1], Weihao Xuan$^{3}$,\\\textbf{Richard Tong}$^{4}$, \textbf{Mona Diab}$^{1}$, \textbf{Tom Mitchell}$^{1}$ \\
  $^{1}$Carnegie Mellon University, $^{2}$University of Maryland\\ $^{3}$The University of Tokyo, $^{4}$NEOLAF \\
  \texttt{zhihaoyu@andrew.cmu.edu}, \texttt{\{yunzex,mdiab,mitchell\}@cs.cmu.edu}\\ \texttt{minglii@umd.edu}, \texttt{weihaoxuan@g.ecc.u-tokyo.ac.jp}, \texttt{richard.tong@ieee.org}
}

\begin{document}
\maketitle


\begin{abstract}

Educational innovation is often constrained by the cost, time, and ethical complexity of human-subject studies. Simulated students offer a complementary methodology: learner models that support instructional testing, data generation, teacher training, and social learning  \citep{vanlehn1998applications, koedinger2015methods}. Large language models (LLMs) substantially expand the expressive power of such simulators, but also introduce a validity risk: fluent interaction can obscure unrealistic error patterns and learning dynamics. We argue that this \emph{competence paradox} arises when a broadly capable model is asked to behave as a partially knowledgeable learner. We reframe the problem as one of \emph{epistemic specification}: educational validity depends on explicitly defining what the simulated learner can access at a given moment (knowledge, strategies, representations, and resources), how errors are structured, and whether and how the state evolves with instruction. We formalize LLM-based student simulation as \emph{constrained generation} under such specifications and propose a goal-by-environment framework that situates simulators by pedagogical goals and deployment contexts. We conclude with open challenges for building valid, reusable, and safe LLM-based student simulators.
\end{abstract}

\section{Introduction}
Educational innovation is often constrained by the logistical, ethical, and financial challenges of human-subject research. Studying and deploying new educational interventions requires navigating informed-consent obligations, mitigating privacy risks in digital settings, and managing non-random attrition, all while bearing the costs of implementation across diverse sites \citep{pardo2014ethical, fixsen2005implementation}.

To complement these constraints, researchers have long studied \emph{simulated students}: computational models of learners that can be instantiated in diverse pedagogical roles. Classic work emphasizes applications including teacher training, learning-by-teaching and collaboration, and formative use in instructional development \citep{vanlehn1998applications}. 

While student simulation has a long history within intelligent tutoring systems, recent work has explored the use of Large Language Models (LLMs) as a flexible substrate for simulated students. By supporting open-ended natural-language interaction, LLM-based systems lower the barrier to constructing interactive learner models in less structured domains and interaction settings \citep{chu-etal-2025-llm}. This flexibility has motivated investigations into richer forms of simulated behavior, including extended interaction over time and participation in multi-agent learning environments \citep{zhang-etal-2025-simulating}.

At the same time, increased fluency and generative capacity introduce new threats to educational validity. Convincing language can obscure unrealistic learning dynamics, compress learner heterogeneity into stereotypes, or introduce artifacts that mislead interpretation \citep{li2025llm}. A key failure mode is the \textit{competence paradox}: models trained to be broadly capable are tasked with exhibiting partial knowledge and misconception-driven behavior. We argue that this tension reflects a mismatch between model capability and the learner state the simulator is intended to represent. Because LLMs do not inherently bind generation to an explicit learner state, their errors can resemble superficial deviations from expert reasoning rather than stable, diagnosis-relevant misconceptions. As a result, educational validity depends on explicitly specifying what the simulated learner can and cannot access, and how its behavior is expected to change under instruction.

\begin{tcolorbox}[
    enhanced,
    colframe=black!50,
    colback=black!2,
    coltitle=white,
    fonttitle=\bfseries\large,
    title={Student Simulation},
    boxrule=0pt,
    leftrule=1.2mm,
    arc=2mm,
    left=8pt,
    right=8pt,
    top=6pt,
    bottom=6pt
]
\large
\textbf{Student simulation} refers to generative systems that produce learner-like behaviors
(e.g., answers, questions, errors, or dialogue moves).
We distinguish student simulation from \emph{student modeling}, which focuses on inferring or
predicting learner states without enacting interactive behavior.
We also note that contemporary LLM-based student simulators are rarely used to test pedagogical
theories; accordingly, we focus on LLM-based simulated students in current practice.
\label{dataexample:kkpuzzle}
\end{tcolorbox}

To make educational claims comparable and evaluations aligned, we call for an explicit \textbf{Epistemic State Specification} as a required system declaration. This specification states what concepts, strategies, representations, and resources are accessible to the simulated learner at a given moment, what gaps or misconceptions structure its behavior, and how this access evolves over time. We situate this requirement within a context-aware framework that organizes simulated student systems along two dimensions: \textit{behavioral goals} (performance, learning, and human aspects) and the \textit{environment} (subject domain, learner population, and interaction modality). Treated as a cross-cutting requirement, the Epistemic State Specification anchors evaluation, prevents overclaiming, and supports meaningful comparison beyond fluency-based proxies.

Our contributions are as follows.
\begin{itemize}[nosep, topsep=0pt]
    \item We position simulated students as a complementary methodology for educational research and system development, spanning roles such as teacher training, data generation, social learning and content evaluation.
    \item We analyze the competence paradox as a mismatch between model capability and intended learner-state access, and formalize student simulation as constrained generation under an explicit Epistemic State Specification.
    \item We propose a context-aware framework that categorizes simulated student systems by behavioral goals and environment to support evaluation alignment and comparison.
    \item We identify open challenges for cumulative progress, including validity threats, harm risks, and the need for reusable baselines and standardized evaluations.
\end{itemize}

\section{Related Works}



\subsection{Traditional Student Simulation}
Simulated students have long served as a bridge between learning theory and the design and evaluation of educational systems, supporting applications such as teacher training, learning-by-teaching, instructional evaluation, and content authoring \citep{vanlehn1998applications, koedinger2015methods}. 

Early work was dominated by rule-based cognitive models that adopt a white-box view of learning and error, where behavior arises from explicit symbolic mechanisms. Errors were modeled as systematic procedural bugs \citep{brown1982toward}, and later formalized within cognitive architectures such as ACT-R \citep{anderson1995cognitive, taatgen2010past}. Systems like SimStudent and the Apprentice Learner extended this paradigm by learning production rules from interaction while retaining explicit representations that support diagnosis and pedagogical intervention \citep{matsuda2007simstudent, maclellan2016apprentice}. Although theoretically grounded and interpretable, these approaches require substantial manual structure and scale poorly to open-ended domains.


\subsection{LLM-Based Student Simulation}
Recent work leverages large language models (LLMs) as simulated students, exploiting their low authoring cost and flexible dialogue. In learning-by-teaching settings, prior systems constrain LLM agents with explicit or evolving learner states to regulate competence and promote productive interaction, demonstrating that educational usefulness depends on epistemic control rather than fluency alone \citep{jin2024teach, rogers2025playing}. 

A parallel line of work scales from individual learners to classroom-level simulations, using role grounding, planning, and memory to sustain multi-agent educational interactions \citep{yue2024mathvc, zhang-etal-2025-simulating}. Related efforts seek to elicit structured reasoning traces and misconceptions from LLM-based learners to support interpretability and formative use \citep{sonkar2024malalgopy}. Together, these studies highlight both the promise and the risk of LLM-based simulators: while expressive and scalable, their validity hinges on explicit specification of learner state and learning dynamics.
\section{Formalizing and Validating LLM-based Simulated Students}
Large language models (LLMs) have enabled a recent shift in educational simulation research from modeling expert tutors toward modeling learners themselves. Rather than focusing on the optimization of instructional policies alone, LLM-based systems increasingly aim to reproduce learner behavior \cite{QI2026130753}, reasoning trajectories \cite{genstu}, and interactional dynamics \cite{zhou2025socialworldmodels}. This shift is not a superficial role-play change. A simulated student must be intentionally bounded: unlike a general-purpose assistant optimized for helpfulness and correctness, a high-fidelity simulated student must reproduce the characteristic imperfections of human learning, including systematic misconceptions, non-optimal reasoning paths, affect-driven behaviors (e.g., anxiety, disengagement), and learning trajectories that evolve over time.

\subsection{Student simulation as constrained generation.}
A core challenge is the \textit{competence paradox}, which we argue is rooted in an irreducible \textit{prior-knowledge entanglement} problem. Mainstream LLMs are trained to be broadly capable, self-correcting, and prosocial. Unlike human learners, they cannot genuinely ``unknow'' solution schemas or expert heuristics once internalized, so even when prompted to act like a novice, their latent reasoning trajectories remain shaped by expert priors that the target student has not acquired.

We therefore define LLM-based student simulation as a \textbf{constrained generation task} whose goal is not to maximize correctness, but to generate responses that remain within an explicit \textit{epistemic boundary}, namely a specification of what the simulated learner can legitimately access at a given moment (concepts, strategies, representations, and resources), together with the misconceptions or gaps that structure their errors. 

Concretely, a simulated student must satisfy three coupled requirements:
\begin{enumerate}[nosep]
    \item \textit{Fidelity of Error}: given a target misconception (or limitation), the simulator should apply it consistently on the problem types where it is relevant;
    \item \textit{Epistemic Consistency}: generated mistakes and explanations should be causally attributable to the stated epistemic boundary, and remain stable across isomorphic items and over multi-turn interaction, rather than appearing as one-off surface deviations from expert reasoning; and
    \item \textit{Boundary of Competence}: outside those regions, the simulator should still behave according to its assumed ability level, avoiding both expert shortcuts that leak inaccessible knowledge and degeneration into random noise.
\end{enumerate}

This framing motivates explicit control mechanisms (prompting, decoding constraints, state variables, external controllers, or structured knowledge representations) that counter the model's default tendency to self-correct and to converge on globally optimal reasoning, while allowing the epistemic boundary to be dynamically updated as the simulated learner progresses.

\subsection{Designing Student Simulation}

To make student simulators comparable and to align evaluation with what is actually being claimed, we organize the design space along two fundamental dimensions: \textit{Behavioral Goals} and \textit{Environment}. \textit{Behavioral goals} specify \emph{which} aspects of learner behavior a simulator aims to reproduce (e.g., response distributions, learning trajectories, or socio-affective behaviors). \textit{Environment} specifies \emph{where} these behaviors are expressed and constrained, including the subject domain, the target learner population, and the interaction modality. The key implication is that two systems can both be called ``simulated students'' yet be incommensurate: a simulator designed to match error distributions in short-answer math under objective grading is not directly comparable to one designed to model long-horizon learning and help-seeking in open-ended dialogue. Making these two dimensions explicit prevents overclaiming and enables evaluations that test the right notion of fidelity for a given setting.

\subsubsection{Behavioral Goals}
Behavioral goals specify which facets of learner behavior the simulator must replicate. We distinguish three goals that are often conflated in prior work.

\paragraph{Simulating Performance.}
The simulator reproduces a student's observable outputs at a time point, including success rates and characteristic error patterns. This goal is central for question difficulty estimation, distractor generation, and synthetic data augmentation where realistic response distributions matter more than modeling how the student acquired or failed to acquire the underlying knowledge.

\paragraph{Simulating Learning.}
The simulator models the \textit{process} of learning as a trajectory. Unlike static performance snapshots, learning simulation requires state evolution across interactions, including skill acquisition, forgetting, and differential sensitivity to interventions (e.g., scaffolding, feedback timing, peer effects). This goal is essential for teacher training (observing pedagogical consequences) and for training adaptive tutors, where policies depend on how the student changes rather than what the student answers once.

\paragraph{Simulating Human Aspects.}
The simulator reproduces non-cognitive attributes that shape learning interactions, such as personality, motivation, emotion, and socio-linguistic style. These factors determine whether classroom and peer-learning interactions feel behaviorally plausible. Affective realism also requires modeling help-seeking behaviors (instrumental, executive, avoidant) that vary with confidence, anxiety, and perceived social risk, rather than defaulting to uniformly cooperative ``good student'' behavior.

\subsubsection{Environment}
Environment specifies external constraints that bound what realism means in practice, and what inputs and modalities are available.

\paragraph{Subjects.}
Different disciplines require different knowledge representations and error models. Structured domains (e.g., math, programming) afford objective correctness and well-defined misconception patterns; open-ended domains (e.g., writing, discussion) require modeling subjective reasoning, argumentation strategies, and rubric-dependent evaluation.

\paragraph{Student group and persona consistency.}
A simulator should match the target population (age, proficiency, language background, cultural context, neurodiversity) because these factors shape both misconceptions and interaction style. Persona is not only a prompt attribute. It should function as a consistency constraint across turns and tasks, preventing implausible ability drift (e.g., claiming low attention while exhibiting sustained high-focus reasoning).

\paragraph{Learning environment and modality.}
Interaction setting constrains observable behavior. Text-only dialogue differs from classrooms or VR settings where speech, gaze, and embodied actions matter. Even within text, the availability of artifacts (equations, diagrams, code editors) changes what errors look like. Environment modeling should also account for instructional structure, such as scaffolding and fading: a realistic student may rely on prompts and deteriorate when supports are removed, rather than instantly generalizing.

\subsection{Epistemic State Specification}
In addition to the two preceding dimensions, we require each simulated-student system to explicitly declare its \textit{epistemic state specification} (ESS). ESS defines \emph{what the learner knows and can access at a given moment}, \emph{how errors are generated}, and \emph{whether and how that state changes over time}. Concretely, an ESS specifies (i) the representations, knowledge elements, strategies, and resources available to the learner at time $t$; (ii) the sources of systematic error, such as misconceptions or incomplete procedures; and (iii) the update mechanism, if any, governing transitions between states.

ESS is intended to clarify the degree to which a system simulates \emph{performance at a fixed competence level} versus \emph{learning as a stateful process}. We operationalize ESS as a lightweight reporting label with five levels:

\begin{itemize}
    \item \textbf{E0: Unspecified}. The learner’s internal knowledge, error sources, and state transitions are not defined. Outputs are generated without an explicit epistemic constraint.
    
    \item \textbf{E1: Static bounded}. The learner operates with a fixed, pre-specified set of knowledge elements, skills, or error templates that do not change during interaction. Behavior reflects performance given an initial competence level, without learning.
    
    \item \textbf{E2: Curriculum-indexed}. The learner’s accessible knowledge or error patterns are updated according to an external progression signal, such as curriculum position, problem index, or mastery variable, without an explicit model of misconceptions or strategy change.
    
    \item \textbf{E3: Misconception-structured}. The learner is governed by an explicit, stable model of misconceptions, strategies, or partial procedures that causally determine behavior. Errors arise from identifiable epistemic structures rather than surface randomness.
    
    \item \textbf{E4: Calibrated or learned}. The learner’s state representation and transition dynamics are learned from or calibrated against human interaction data, such that both performance and state evolution are empirically grounded in observed learner trajectories.
\end{itemize}

This label functions as a cross-cutting declaration intended to prevent overclaiming, enable meaningful comparison across systems, and align evaluation protocols with the simulator’s stated epistemic constraints. Illustrative examples of several ESS level are provided in Appendix~\ref{sec:case_studies}.

\section{Promising Directions of LLM-based Student Simulation} 

This section outlines where LLM-based simulated students are most promising and why. We organize the opportunities into four directions: \textit{Teacher Training}, \textit{Social Learning} (including Learning-by-Teaching and Collaborative Learning), \textit{Data Generation}, and \textit{Content Evaluation}. For each direction, we highlight the educational motivation and distinctive capabilities that LLMs bring. \textbf{Table \ref{tab:context_aware_framework} summarizes these four directions alongside three unifying benefits that LLM-based simulated students provide across all of them:} \textbf{Scalability} (deployment beyond human availability), \textbf{Safety} (risk-free experimentation that reduces privacy, liability, and ethical concerns), and \textbf{Versatility} (behavioral modeling of diverse learner characteristics that are difficult or impossible to reproduce reliably with real students). 
\begin{table*}[t]
\centering
\caption{Dimensional Mapping of Simulated Student Research: Goals and Environment}
\label{tab:context_aware_framework}
\small
\renewcommand{\arraystretch}{1.5}
\begin{tabular}{|p{3cm}|p{5.5cm}|p{5.5cm}|}
\hline
\textbf{Applied Context} & \textbf{Associated Behavioral Goals} & \textbf{Environmental Considerations} \\ \hline
\textbf{Teacher Training} & High fidelity in \textit{Simulating Learning} (trajectories) and \textit{Human Aspects} (affect, personality). & Requires high student group heterogeneity and classroom-like interaction settings, potentially multimodal. \\ \hline
\textbf{Social Learning} & Focus on \textit{Human Aspects} (socio-linguistics) and \textit{Simulating Learning} (to induce prot\'eg\'e effect). & Dialogue-based interfaces; collaborative or 1-on-1 peer environments. \\ \hline
\textbf{Data Generation} & Emphasis on \textit{Simulating Performance} (error patterns) and \textit{Learning} (longitudinal logs). & Often structured subjects (STEM, programming); digital interaction logs and standardized tasks. \\ \hline
\textbf{Content Evaluation} & Primary focus on \textit{Simulating Performance} (static success/failure distributions). & High-throughput platform integration; objective correctness criteria and scalable test suites. \\ \hline
\end{tabular}
\end{table*}
\subsection{Teacher Training}
High-quality instruction is a primary driver of student achievement, whereas inadequate teacher preparation leads to detrimental short- and long-term academic outcomes  \citep{markel-etal-2023-gpteach}. However, providing rigorous training has become increasingly challenging due to the rapid expansion of online education and the reliance on large cohorts of novice tutors and teaching assistants  \citep{pan2025tutorup, markel-etal-2023-gpteach}. Opportunities for actual teaching are often scarce and constrained by factors outside the training program's control  \citep{zheng2023educasim}. Other traditional training solutions, ranging from webinars to human-authored visual or VR simulations, are resource-intensive and time-consuming to maintain.  Consequently, these methods are difficult to deploy at the scale required to standardize training for the growing workforce of pre-service and part-time educators  \citep{lee2023generativeagent, christensen2011simschool}.

Beyond scalability, traditional field experiences are fraught with inherent structural limitations. Training in real classrooms introduces significant accountability, liability, and privacy risks  \citep{christensen2011simschool}. Crucially, limited contact hours restrict the exposure of a novice to student heterogeneity; real-world placements cannot guaranty the variety of personalities, learning styles, and aptitudes necessary to master adaptive instruction  \citep{sanyal2025pedagogical, knezek2024specialneeds}. This lack of diverse exposure hinders the development of complex skills, such as managing whole-class dynamics or facilitating discussion-based on argumentation  \citep{nazaretsky2023ai}.

LLM-based simulated students offer a solution to these challenges by providing frequent, structured, and risk-free rehearsal opportunities  \citep{judge2013visual}. Unlike static environments, LLMs can model diverse, personality-aligned behaviors, creating a rich testbed for practicing adaptive strategies  \citep{ma2025soei, pentangelo2025senemai}. Research indicates that these simulated environments alleviate the immediate time pressure of real classrooms, allowing teachers to draft more thoughtful, inclusive responses and strategize around learning goals  \citep{markel-etal-2023-gpteach}. By embedding these agents in realistic settings, systems like TutorUp enable novices to practice navigating disengagement and confusion in a safe, controllable environment, bridging the gap between theory and practice  \citep{pan2025tutorup, lee2023generativeagent}.

\subsection{Social Learning}
Simulated students can further facilitate education by functioning as social learning partners. Learning by Teaching (LbT) is a robust pedagogical model where students achieve deeper conceptual understanding through the acts of structuring knowledge for explanation, taking responsibility for a tutee, and reflecting on their own learning processes  \citep{biswas2005learning}. Despite its efficacy across age groups and domains, designing effective LbT environments presents challenges. A critical limitation occurs when neither the tutor nor the tutee is an expert; without guidance, participants often reinforce shared misconceptions or focus on trivial details rather than core concepts  \citep{Debbané:2023:LBT:CSCW}. While human experts can mitigate this by feigning ignorance to elicit explanations, such role-playing is labor-intensive and inherently unscalable  \citep{duran2017learning}.

LLM-based simulated students offer a scalable alternative to this dynamic. These agents can simulate a novice learner to induce the prot\'eg\'e effect, in which students make greater effort to learn material they expect to teach, while simultaneously leveraging their underlying expert knowledge to subtly guide the student toward correct understanding  \citep{rogers2025playing}. Furthermore, teaching an artificial agent creates a psychologically safe environment that mitigates the fear of judgment and the social pressure often associated with real-time human interaction  \citep{Chase:2009:TAP:Journal,Debbané:2023:LBT:CSCW}.

Beyond the tutor--tutee dynamic, simulated students can also serve as peers within collaborative learning frameworks. Positive peer interactions are known to enhance academic outcomes and foster essential soft skills, such as communication and group coordination  \citep{Veldman:2020:YCW:LI,Rohrbeck:2003:PAL:JEP}. Conversely, educational settings that lack real-time feedback and interaction frequently yield suboptimal learning outcomes. In the current digital landscape, which is characterized by pre-recorded lectures, remote instruction, and asynchronous schedules, students often face logistical and social barriers to finding study partners  \citep{Wang:2025:GCL:GROUP}. Simulated students bridge this gap by serving as always-available virtual peers or learning companions that restore the benefits of cooperative interaction to otherwise isolated learning environments.

\subsection{Data Generation}
The efficacy of modern intelligent education platforms (e.g., Coursera) and LLM-based educational tools (e.g., Google LearnLM) relies heavily on the availability of high-quality training data to support personalization and algorithm optimization  \citep{Gao:2025:A4E:arXiv, Song2024LearnLM}. However, the acquisition of high-fidelity educational datasets, such as student responses and conversation logs, is severely constrained by strict privacy regulations and the high costs associated with manual annotation. Furthermore, real-world educational data is often sparse or noisy, while achieving strong algorithmic performance requires comprehensive and well-structured datasets  \citep{Chen:2025:SIM:arXiv, Gao:2025:A4E:arXiv}.

Simulated students address this scarcity by serving as scalable engines for synthetic data generation  \cite{tutorial}. By modeling diverse learner profiles, these agents can produce large-scale, privacy-compliant datasets that mimic real-world distributions. Recent literature demonstrates the utility of this approach in two key areas: generating realistic static student responses  \citep{Miroyan:2025:PGS:arXiv, Benedetto:2024:ULMS, Wu:2025:EISS, Ross:2025:LMM}, and synthesizing complex, longitudinal task-solving trajectories  \citep{Ross:2025:MSL, Sharma:2024:DSS:EDM}.

\subsection{Content Evaluation}
The rapid expansion of educational technology platforms and pedagogical conversational agents (PCAs), such as Khan Academy's Khanmigo, Squirrel AI, and Duolingo, presents both opportunities for personalized instruction and challenges in ensuring content quality and pedagogical efficacy  \citep{Gonnermann-Muller:2025:FCE:arXiv, Jin:2024:TTR:TR}. Evaluating these systems faces critical bottlenecks. Expert teacher assessment is prohibitively expensive, while pilot testing with real students introduces privacy risks and the potential exposure to harmful content  \citep{Ezzaki:2024:EEC:LLM, Jin:2024:TTR:TR}. Simulated students offer a compelling alternative by providing a high-throughput and risk-free environment for both system-level validation and content-level assessment, particularly for item difficulty modeling  \citep{hambleton1991fundamentals, hsu2018automated, alkhuzaey2021systematic, li2025item, peters2025text}. Although accurate item difficulty modeling is essential for assessment validity and adaptive learning algorithms  \citep{Duenas:2024:UPN:BEA}, traditional approaches, including manual expert labeling and item response theory, suffer from subjectivity, limited scalability, or reliance on extensive historical student data. These limitations make simulated students an increasingly attractive solution.
\looseness-1

Simulated students address these limitations by acting as synthetic test-takers. By modeling learners with varying ability levels, these agents can generate performance data to predict item difficulty without the need for large-scale human pilots. Recent frameworks such as UPN-ICC, QG-SMS, and SMART demonstrate that simulated students, whether used independently or to augment IRT models, can achieve high accuracy in calibrating educational content  \citep{Duenas:2024:UPN:BEA, Nguyen:2025:QGS:arXiv, Scarlatos:2025:SMA:arXiv}. At the same time, recent large-scale evidence suggests this setting is not solved by simply substituting an LLM for human pilots: even with difficulty labels grounded in real student field testing, off-the-shelf LLMs can be systematically misaligned with human difficulty judgments, and they struggle to reliably simulate lower-proficiency cognitive states  \citep{Li2025DifficultyAlignment}. This underscores the need for calibrated, goal-conditioned simulated students whose proficiency controls and validation protocols are explicitly specified.
\section{Challenges}
\subsection{Data Scarcity and Privacy}
\label{sec:data-privacy}
High-fidelity student simulation depends on granular, real-world traces that capture learner errors, feedback, and instructional context. In practice, such data are scarce: collecting them is costly (often requiring expert annotation) and sustained access to classrooms or proprietary platforms is difficult to obtain. Missing instructional materials and intervention context can also create a persistent contextual void, limiting a simulator's ability to represent how pedagogy shapes learning outcomes \citep{Zhai:2020:AIReview,Baker:2014:EDM,Xu:2024:EGSA}.

Privacy constraints further restrict both what can be collected and what can be shared. Educational traces frequently contain rich demographic and behavioral identifiers, so many datasets cannot be redistributed, undermining reproducibility and slowing cumulative progress \citep{Hoel:2018:Privacy,Wu:2025:EISS}. For LLM-based simulation, the risk compounds because models trained on educational traces may memorize and reproduce private information, or exhibit privacy-related biases that distort pedagogical interactions \citep{shvartzshnaider2025privacybiaslanguagemodels}.

\subsection{Evaluation and Standardization under Non-Verifiable Outcomes}
As simulated students are increasingly deployed in long-form inquiry, open-ended reasoning, and authentic classroom dialogue, evaluation becomes difficult because the target behavior is rarely verifiable by binary correctness. Recent evidence suggests a broader \textbf{non-verifiability crisis}: standard automated metrics and even LLM-based judges can misalign with expert human preferences, with reported preference-matching accuracy on the order of $\sim$65\% in some scientific reasoning settings \citep{Cohan2025SciArena}. This limitation is amplified by a style-substance mismatch, where models produce responses that are polished and confident yet ungrounded or logically inconsistent. Consequently, automated evaluators may fail to distinguish a student who is productively struggling from a simulator that is convincingly hallucinating \citep{Zhang2025SimulRAG}.

These measurement challenges directly motivate the need for \textbf{standardized evaluation frameworks} that are explicitly aligned with a simulator's intended behavioral goal and deployment environment. In particular, assessing learning dynamics requires different instruments than those used for static correctness. Yet the absence of shared benchmarks and reporting conventions has pushed current work toward ad-hoc expert ratings that are costly to scale, difficult to reproduce, and hard to compare across studies. This ambiguity makes cross-paper comparisons fragile, since the appropriate notion of fidelity depends on the simulator’s behavioral goal and deployment environment, and remains under-specified without shared, goal-conditioned benchmarks.

\section{Recommendation}
\subsection{Mandate Epistemic State Specification for Reproducibility}
Epistemic State Specification (ESS; E0--E4) should be treated as a required reporting artifact for simulated student systems because the competence paradox is ultimately an epistemic mismatch: LLMs cannot truly “unknow,” so they may leak expert knowledge while producing novice-like errors. System descriptions should therefore explicitly state what knowledge, strategies, and resources are accessible at each turn and how this access evolves over time; moving from unspecified (E0) to misconception-structured (E3) or calibrated (E4) specifications makes claims falsifiable (for example, stable misconception behavior under paraphrase and across isomorphic items), turns “student level” into an auditable design choice, and enables meaningful cross-paper comparison.

\subsection{Shift Evaluation from Surface Realism to Goal-Aligned Fidelity}

Evaluation should prioritize goal-aligned fidelity rather than surface realism. The fidelity-evaluation gap arises when fluent dialogue and generic “humanness” scores are taken as evidence of educational validity, even though they can mask implausible learning dynamics. Moreover, simulators built for different purposes (for example data generation versus teacher training) require fundamentally different success criteria.

Metrics should be derived from the simulator’s behavioral goal and environment. Performance and data-generation settings call for controlled error distributions and resistance to drifting into expert reasoning. Learning-oriented settings require trajectory metrics such as gradual improvement and appropriate sensitivity to feedback. Human-aspects and social learning settings require interactional metrics that match the application, such as persistence, help-seeking, or affective coherence when relevant.

\subsection{Integrate Explicit Learning Mechanisms to Model Trajectories}

Longitudinal simulation benefits from explicit learning mechanisms rather than context-window prompting alone. Prompt-only approaches often yield unrealistic dynamics, including abrupt novice-to-expert jumps, unstable competence across turns, and brittle dependence on phrasing, which undermines the validity of using simulators to study learning or stress-test instructional interventions.

A practical direction is hybrid architectures that pair an LLM with an explicit learner-state representation and a defined transition rule, such as knowledge tracing, proficiency variables, misconception graphs, or cognitive models. This makes learning trajectories interpretable and calibratable, and it makes failures diagnosable as state, transition, or interface errors rather than opaque model variance.

\subsection{Establish Standardized Misconception Benchmarks to Address Non-Verifiability}

Shared benchmarks are needed for non-verifiable educational behaviors, especially misconception consistency and longitudinal socio-affective trajectories. Many properties central to educational validity cannot be assessed by correctness alone, and automated judges can be misled by fluent but incoherent behavior, leaving the field reliant on expensive, ad hoc expert ratings and fragile comparisons.

Benchmark suites should emphasize consistency under controlled variation. Misconception tests can use isomorphic items and paraphrases to assess stability of error signatures aligned with declared epistemic states. Learning tests can use multi-turn curricula to assess gradual, selective revision under feedback. Socio-affective tests can use scenarios that elicit frustration or re-engagement to assess coherent trajectories over time. Open, standardized suites would directly improve reproducibility and comparability.

\section{Conclusion}
In conclusion, to advance LLM-based student simulation from exploratory prototypes to rigorous scientific instruments, the field must resolve the ``competence paradox'' by prioritizing \textbf{epistemic fidelity} over surface realism. We formalized this shift through the \textbf{Goal-by-Environment framework} and the \textbf{Epistemic State Specification (ESS)}, which ensure that simulated behaviors are causally attributable to defined learning states rather than stochastic hallucinations. By adopting these standards, researchers can transform simulated students into trustworthy, reproducible testbeds for educational innovation, bridging the gap between conversational fluency and valid pedagogical modeling.
\section*{Limitations}

This study is primarily theoretical and synthesizes insights from previous literature in educational psychology, learning sciences, and natural language processing. We did not conduct empirical evaluations, classroom deployments, or large-scale psychometric validations of the proposed Epistemic State Specification. As such, our claims regarding the resolution of the ``competence paradox'' are not validated through direct pedagogical interaction or longitudinal system testing. Although we propose a distinct taxonomy for epistemic states (E0--E4), real-world model behaviors often exhibit fluctuating or indeterminate knowledge boundaries. Our framework idealizes these dimensions for analytical clarity, which may limit its robustness when applied to stochastic, non-deterministic foundation models. 

Moreover, the proposed solution assumes that developers can enforce strict knowledge constraints via architectural design, which may be undermined by the inherent fragility of prompt engineering and the opacity of commercial black-box models. Future work should focus on validating the framework through controlled empirical studies that measure whether ESS-compliant simulators effectively prevent knowledge leakage and by developing automated metrics to quantify epistemic fidelity across diverse learning scenarios.

\section*{Ethical Considerations}

The deployment of LLM-based student simulators entails significant pedagogical risks if the generated behaviors are not rigorously grounded in well-defined epistemic states. A primary concern is the potential for negative training transfer in teacher education, in which novice instructors may develop maladaptive teaching strategies by practicing with simulators that exhibit realistic fluency but unrealistic learning dynamics. For example, if a simulator responds to a teaching intervention with plausible yet causally disconnected improvements, such as appearing to master a complex concept immediately after a simple hint, it reinforces superficial instructional moves rather than deep pedagogical reasoning  \cite{dai2025embracingcontradictiontheoreticalinconsistency}. This misalignment creates a hazard in which the simulator functions as a ``pedagogical placebo,'' offering the illusion of effective practice while failing to reflect the cognitive resistance and gradual skill acquisition observed in real classrooms.

\section*{Acknowledgements}
We thank Prof.\ Ken Koedinger and Prof.\ Carolyn Ros\'{e} for their valuable feedback and suggestions which we heavily incoporated into this paper. We also thank Gati Aher and Aylin \"{O}ZT\"{U}RK for providing helpful suggestions during the early-stage ideation of this project.

Furthermore, LLM-based student profiles can establish normative or cultural biases if the behavioral and linguistic cues associated with ``struggle'' or ``misconception'' are not localized or participatory in design \cite{xiao-etal-2025-humanizing}. This marginalizes underrepresented cultures \cite{alkhamissi2025hireanthropologistrethinkingculture} or reinforces dominant stereotypes regarding academic ability. Although our Goal-by-Environment framework advocates for context-sensitive profile calibration, more empirical research is needed to verify the effectiveness of such strategies across diverse educational settings.

We also recognize the potential for dual use of our framework. The taxonomy of ``human aspects'' could be used to inform more persuasive or emotionally manipulative systems, especially in commercial tutoring or surveillance contexts. We encourage future work to develop mitigation strategies, such as interpretability indicators (e.g., exposing the active ESS state to the user), constrained anthropomorphic profiles, or gated release mechanisms, to help monitor and control simulated student behavior. We also stress the importance of interdisciplinary collaboration with learning scientists, ethicists, and affected communities during system development.

By articulating both the functional benefits and the possible harms of epistemic simulation in LLMs, our goal is to support transparent, socially aligned, and user-aware design practices. We strongly encourage future research to empirically validate and refine this framework, particularly through participatory codesign and cross-cultural evaluation.

\bibliography{latex/custom,latex/anthology_0,latex/anthology_1}

\begin{thebibliography}{71}
\providecommand{\natexlab}[1]{#1}

\bibitem[{AlKhamissi et~al.(2025)AlKhamissi, Xiao, AlKhamissi, and Diab}]{alkhamissi2025hireanthropologistrethinkingculture}
Mai AlKhamissi, Yunze Xiao, Badr AlKhamissi, and Mona Diab. 2025.
\newblock \href {https://arxiv.org/abs/2510.05931} {Hire your anthropologist! rethinking culture benchmarks through an anthropological lens}.
\newblock \emph{Preprint}, arXiv:2510.05931.

\bibitem[{AlKhuzaey et~al.(2021)AlKhuzaey, Grasso, Payne, and Tamma}]{alkhuzaey2021systematic}
Samah AlKhuzaey, Floriana Grasso, Terry~R Payne, and Valentina Tamma. 2021.
\newblock A systematic review of data-driven approaches to item difficulty prediction.
\newblock In \emph{International conference on artificial intelligence in education}, pages 29--41. Springer.

\bibitem[{Anderson et~al.(1995)Anderson, Corbett, Koedinger, and Pelletier}]{anderson1995cognitive}
John~R. Anderson, Albert~T. Corbett, Kenneth~R. Koedinger, and Ray Pelletier. 1995.
\newblock \href {https://doi.org/10.1207/s15327809jls0402_2} {Cognitive tutors: Lessons learned}.
\newblock \emph{The Journal of the Learning Sciences}, 4(2):167--207.

\bibitem[{Baker and Siemens(2014)}]{Baker:2014:EDM}
Ryan~S. Baker and George Siemens. 2014.
\newblock Educational data mining and learning analytics.
\newblock \emph{Cambridge Handbook of the Learning Sciences}, pages 253--272.

\bibitem[{Benedetto et~al.(2024)Benedetto, Aradelli, Donvito, Lucchetti, Cappelli, and Buttery}]{Benedetto:2024:ULMS}
Luca Benedetto, Giovanni Aradelli, Antonia Donvito, Alberto Lucchetti, Andrea Cappelli, and Paula Buttery. 2024.
\newblock Using llms to simulate students’ responses to exam questions.
\newblock In \emph{Findings of the Association for Computational Linguistics: EMNLP 2024}, pages 11351--11368.

\bibitem[{Biswas et~al.(2005)Biswas, Leelawong, Schwartz, Vye, and at~Vanderbilt}]{biswas2005learning}
Gautam Biswas, Krittaya Leelawong, Daniel Schwartz, Nancy Vye, and The Teachable Agents~Group at~Vanderbilt. 2005.
\newblock Learning by teaching: A new agent paradigm for educational software.
\newblock \emph{Applied Artificial Intelligence}, 19(3-4):363--392.

\bibitem[{Brown and VanLehn(1982)}]{brown1982toward}
John~Seely Brown and Kurt VanLehn. 1982.
\newblock Toward a generative theory of "bugs".
\newblock In Thomas~P. Carpenter, James~M. Moser, and Thomas~A. Romberg, editors, \emph{Addition and subtraction: A cognitive perspective}, pages 117--136. Erlbaum, Hillsdale, NJ.

\bibitem[{Chase et~al.(2009)Chase, Chin, Oppezzo, and Schwartz}]{Chase:2009:TAP:Journal}
Catherine~C. Chase, Doris~B. Chin, Marily~A. Oppezzo, and Daniel~L. Schwartz. 2009.
\newblock Teachable agents and the prot\'{e}g\'{e} effect: Increasing the effort towards learning.
\newblock \emph{Journal of Science Education and Technology}, 18:334--352.

\bibitem[{Chen et~al.(2025)Chen, Molnar, Hua, Li, Khiem, Ambrose, Lang, Metoyer, and Chawla}]{Chen:2025:SIM:arXiv}
Si~Chen, Izzy Molnar, Ting Hua, Peiyu Li, Le~Huy Khiem, G~Alex Ambrose, Jim Lang, Ronald Metoyer, and Nitesh~V Chawla. 2025.
\newblock $\textsc{SimInstruct}$: A responsible tool for collecting scaffolding dialogues between experts and llm-simulated novices.
\newblock \emph{arXiv preprint arXiv:2508.04428}.

\bibitem[{Christensen et~al.(2011)Christensen, Knezek, Tyler-Wood, and Gibson}]{christensen2011simschool}
Rhonda Christensen, Gerald Knezek, Tandra Tyler-Wood, and David Gibson. 2011.
\newblock Simschool: An online dynamic simulator for enhancing teacher preparation.
\newblock \emph{International Journal of Learning Technology}, 6(2):201--220.

\bibitem[{Chu et~al.(2025)Chu, Wang, Xie, Zhu, Yan, Ye, Zhong, Hu, Liang, Yu, and Wen}]{chu-etal-2025-llm}
Zhendong Chu, Shen Wang, Jian Xie, Tinghui Zhu, Yibo Yan, Jingheng Ye, Aoxiao Zhong, Xuming Hu, Jing Liang, Philip~S. Yu, and Qingsong Wen. 2025.
\newblock \href {https://doi.org/10.18653/v1/2025.findings-emnlp.743} {{LLM} agents for education: Advances and applications}.
\newblock In \emph{Findings of the Association for Computational Linguistics: EMNLP 2025}, pages 13782--13810, Suzhou, China. Association for Computational Linguistics.

\bibitem[{Dai and Xiao(2025)}]{dai2025embracingcontradictiontheoreticalinconsistency}
Gordon Dai and Yunze Xiao. 2025.
\newblock \href {https://arxiv.org/abs/2505.18139} {Embracing contradiction: Theoretical inconsistency will not impede the road of building responsible ai systems}.
\newblock \emph{Preprint}, arXiv:2505.18139.

\bibitem[{Debban\'{e} et~al.(2023)Debban\'{e}, Lee, Tse, and Law}]{Debbané:2023:LBT:CSCW}
Amy Debban\'{e}, Ken~Jen Lee, Jarvis Tse, and Edith Law. 2023.
\newblock \href {https://doi.org/10.1145/3579501} {Learning by teaching: Key challenges and design implications}.
\newblock \emph{Proceedings of the ACM on Human-Computer Interaction (PACM HCI)}, 7(CSCW1):Article 68.

\bibitem[{Dueñas et~al.(2024)Dueñas, Jimenez, and Ferro}]{Duenas:2024:UPN:BEA}
George Dueñas, Sergio Jimenez, and Geral Eduardo~Mateus Ferro. 2024.
\newblock Upn-icc at bea 2024 shared task: Leveraging llms for multiple-choice questions difficulty prediction.
\newblock In \emph{Proceedings of the 19th Workshop on Innovative Use of NLP for Building Educational Applications (BEA 2024)}.

\bibitem[{Duran(2017)}]{duran2017learning}
David Duran. 2017.
\newblock \href {https://doi.org/10.1080/14703297.2015.1110031} {Learning-by-teaching. evidence and implications as a pedagogical mechanism}.
\newblock \emph{Innovations in Education and Teaching International}, 54(5):476--484.

\bibitem[{Ezzaki et~al.(2024)Ezzaki, Messaoudi, and Naciri}]{Ezzaki:2024:EEC:LLM}
Soukaina Ezzaki, Najat Messaoudi, and Jaafar~Khalid Naciri. 2024.
\newblock Evaluating educational content through virtual student simulations using large language models.
\newblock \emph{Evaluation of Educational Content Through Virtual Student Simulations Using Large Language Models}.

\bibitem[{Fixsen et~al.(2005)Fixsen, Naoom, Blase, Friedman, and Wallace}]{fixsen2005implementation}
Dean~L. Fixsen, Sandra~F. Naoom, Karen~A. Blase, Robert~M. Friedman, and Frances Wallace. 2005.
\newblock \href {https://nirn.fpg.unc.edu/resources/implementation-research-synthesis-literature} {Implementation research: A synthesis of the literature}.
\newblock National Implementation Research Network.

\bibitem[{Gao et~al.(2025)Gao, Liu, Yue, Yao, Lv, Zhang, Wang, and Huang}]{Gao:2025:A4E:arXiv}
Weibo Gao, Qi~Liu, Linan Yue, Fangzhou Yao, Rui Lv, Zheng Zhang, Hao Wang, and Zhenya Huang. 2025.
\newblock Agent4edu: Generating learner response data by generative agents for intelligent education systems.
\newblock In \emph{Proceedings of the AAAI Conference on Artificial Intelligence}, volume~39, pages 23923--23932.

\bibitem[{Gonnermann-M{\"u}ller et~al.(2025)Gonnermann-M{\"u}ller, Haase, Fackeldey, and Pokutta}]{Gonnermann-Muller:2025:FCE:arXiv}
Jana Gonnermann-M{\"u}ller, Jennifer Haase, Konstantin Fackeldey, and Sebastian Pokutta. 2025.
\newblock Facet: Teacher-centred llm-based multi-agent systems-towards personalized educational worksheets.
\newblock \emph{arXiv preprint arXiv:2508.11401}.

\bibitem[{Hambleton et~al.(1991)Hambleton, Swaminathan, and Rogers}]{hambleton1991fundamentals}
Ronald~K Hambleton, Hariharan Swaminathan, and H~Jane Rogers. 1991.
\newblock \emph{Fundamentals of item response theory}, volume~2.
\newblock Sage.

\bibitem[{Hoel and Chen(2018)}]{Hoel:2018:Privacy}
Tore Hoel and Weiqin Chen. 2018.
\newblock Privacy and data protection in learning analytics: A stakeholder-centered taxonomy.
\newblock \emph{Research and Practice in Technology Enhanced Learning}, 13(1):1--21.

\bibitem[{Hsu et~al.(2018)Hsu, Lee, Chang, and Sung}]{hsu2018automated}
Fu-Yuan Hsu, Hahn-Ming Lee, Tao-Hsing Chang, and Yao-Ting Sung. 2018.
\newblock Automated estimation of item difficulty for multiple-choice tests: An application of word embedding techniques.
\newblock \emph{Information Processing \& Management}, 54(6):969--984.

\bibitem[{Jin et~al.(2024)Jin, Lee, Shin, and Kim}]{jin2024teach}
Hyoungwook Jin, Seonghee Lee, Hyungyu Shin, and Juho Kim. 2024.
\newblock \href {https://doi.org/10.1145/3613904.3642349} {Teach {AI} how to code: Using large language models as teachable agents for programming education}.
\newblock In \emph{Proceedings of the {CHI} Conference on Human Factors in Computing Systems (CHI '24)}, Honolulu, HI, USA. Association for Computing Machinery.

\bibitem[{Jin et~al.(2025)Jin, Yoo, Park, Lee, Wang, and Kim}]{Jin:2024:TTR:TR}
Hyoungwook Jin, Minju Yoo, Jeongeon Park, Yokyung Lee, Xu~Wang, and Juho Kim. 2025.
\newblock Teachtune: Reviewing pedagogical agents against diverse student profiles with simulated students.
\newblock In \emph{Proceedings of the 2025 CHI Conference on Human Factors in Computing Systems}, pages 1--28.

\bibitem[{Judge et~al.(2013)Judge, Bobzien, Maydosz, Gear, and Katsioloudis}]{judge2013visual}
Sharon Judge, Jonna Bobzien, Ann Maydosz, Sabra Gear, and Petros Katsioloudis. 2013.
\newblock The use of visual-based simulated environments in teacher preparation.
\newblock \emph{Journal of Education and Training Studies}, 1(1).

\bibitem[{K{\"a}ser and Alexandron(2024)}]{kaser-alexandron-2024-simulated}
Tanja K{\"a}ser and Giora Alexandron. 2024.
\newblock Simulated learners in educational technology: A systematic literature review and a turing-like test.
\newblock \emph{International Journal of Artificial Intelligence in Education}, 34(2):545--585.

\bibitem[{Knezek and Christensen(2024)}]{knezek2024specialneeds}
Gerald Knezek and Rhonda Christensen. 2024.
\newblock Improving teaching effectiveness for students with special learning needs through on-line simulations.
\newblock In \emph{International Journal on E-Learning}, pages 133--152. Association for the Advancement of Computing in Education (AACE).

\bibitem[{Koedinger et~al.(2015)Koedinger, Matsuda, MacLellan, and McLaughlin}]{koedinger2015methods}
Kenneth~R Koedinger, Noboru Matsuda, Christopher~J MacLellan, and Elizabeth~A McLaughlin. 2015.
\newblock Methods for evaluating simulated learners: Examples from simstudent.
\newblock In \emph{AIED Workshops}.

\bibitem[{Lee et~al.(2023)Lee, Lee, Koh, Jeong, Jung, Byun, Lee, Moon, Lim, and Kim}]{lee2023generativeagent}
Unggi Lee, Sanghyeok Lee, Junbo Koh, Yeil Jeong, Haewon Jung, Gyuri Byun, Yunseo Lee, Jewoong Moon, Jieun Lim, and Hyeoncheol Kim. 2023.
\newblock Generative agent for teacher training: Designing educational problem-solving simulations with large language model-based agents for pre-service teachers.
\newblock In \emph{NeurIPS’23 Workshop on Generative AI for Education (GAIED)}.

\bibitem[{Li et~al.(2025{\natexlab{a}})Li, Chen, Namkoong, and Peng}]{li2025llm}
Ang Li, Haozhe Chen, Hongseok Namkoong, and Tianyi Peng. 2025{\natexlab{a}}.
\newblock Llm generated persona is a promise with a catch.
\newblock \emph{arXiv preprint arXiv:2503.16527}.

\bibitem[{Li et~al.(2025{\natexlab{b}})Li, Huang, Li, Zhou, Zhang, and Liu}]{tutorial}
Dawei Li, Yue Huang, Ming Li, Tianyi Zhou, Xiangliang Zhang, and Huan Liu. 2025{\natexlab{b}}.
\newblock \href {https://doi.org/10.1145/3746252.3761455} {Generative models for synthetic data: Transforming data mining in the genai era}.
\newblock In \emph{Proceedings of the 34th ACM International Conference on Information and Knowledge Management}, CIKM '25, page 6833–6836, New York, NY, USA. Association for Computing Machinery.

\bibitem[{Li et~al.(2025{\natexlab{c}})Li, Chen, Xiao, Chen, Jiao, and Zhou}]{Li2025DifficultyAlignment}
Ming Li, Han Chen, Yunze Xiao, Jian Chen, Hong Jiao, and Tianyi Zhou. 2025{\natexlab{c}}.
\newblock \href {https://arxiv.org/abs/2512.18880} {Can {LLM}s estimate student struggles? human-ai difficulty alignment with proficiency simulation for item difficulty prediction}.
\newblock \emph{arXiv preprint arXiv:2512.18880}.

\bibitem[{Li et~al.(2025{\natexlab{d}})Li, Jiao, Zhou, Zhang, Peters, and Lissitz}]{li2025item}
Ming Li, Hong Jiao, Tianyi Zhou, Nan Zhang, Sydney Peters, and Robert~W Lissitz. 2025{\natexlab{d}}.
\newblock Item difficulty modeling using fine-tuned small and large language models.
\newblock \emph{Educational and Psychological Measurement}, page 00131644251344973.

\bibitem[{Lu and Wang(2024)}]{genstu}
Xinyi Lu and Xu~Wang. 2024.
\newblock \href {https://doi.org/10.1145/3657604.3662031} {Generative students: Using llm-simulated student profiles to support question item evaluation}.
\newblock In \emph{Proceedings of the Eleventh ACM Conference on Learning @ Scale}, L@S '24, page 16–27, New York, NY, USA. Association for Computing Machinery.

\bibitem[{Ma et~al.(2024)Ma, Shen, Koedinger, and Wu}]{ma-etal-2024-teach}
Qianou Ma, Hua Shen, Kenneth Koedinger, and Tongshuang Wu. 2024.
\newblock \href {https://doi.org/10.1007/978-3-031-64302-6_1} {How to teach programming in the ai era? using llms as a teachable agent for debugging}.
\newblock In \emph{Proceedings of the 25th International Conference on Artificial Intelligence in Education (AIED 2024)}, volume 14829 of \emph{Lecture Notes in Artificial Intelligence}, pages 1--16, Recife, Brazil. Springer.

\bibitem[{Ma et~al.(2025)Ma, Hu, Li, Wang, Liu, Cheong, and Chen}]{ma2025soei}
Yiping Ma, Shiyu Hu, Xuchen Li, Yipei Wang, Shiqing Liu, Kang~Hao Cheong, and Yuqing Chen. 2025.
\newblock When llms learn to be students: The soei framework for modeling and evaluating virtual student agents in educational interaction.
\newblock ArXiv preprint arXiv:2410.15701v2.

\bibitem[{MacLellan et~al.(2016)MacLellan, Harpstead, Patel, and Koedinger}]{maclellan2016apprentice}
Christopher~J. MacLellan, Erik Harpstead, Rony Patel, and Kenneth~R. Koedinger. 2016.
\newblock The apprentice learner architecture: Closing the loop between learning theory and educational data.
\newblock In \emph{Proceedings of the 9th International Conference on Educational Data Mining (EDM)}, pages 151--158.

\bibitem[{Markel et~al.(2023)Markel, Opferman, Landay, and Piech}]{markel-etal-2023-gpteach}
Julia~M. Markel, Steven~G. Opferman, James~A. Landay, and Chris Piech. 2023.
\newblock \href {https://doi.org/10.1145/3573051.3593393} {{GPT}each: Interactive {TA} training with {GPT}-based students}.
\newblock In \emph{Proceedings of the Tenth ACM Conference on Learning @ Scale (L@S '23)}, pages 1--11, Copenhagen, Denmark. Association for Computing Machinery.

\bibitem[{Matsuda et~al.(2007)Matsuda, Cohen, Sewall, Lacerda, and Koedinger}]{matsuda2007simstudent}
Noboru Matsuda, William~W. Cohen, Jonathan Sewall, Gustavo Lacerda, and Kenneth~R. Koedinger. 2007.
\newblock Simstudent: Building an intelligent tutoring system by tutoring a synthetic student.
\newblock Submitted manuscript.

\bibitem[{Miroyan et~al.(2025)Miroyan, Niousha, Gonzalez, Ranade, and Norouzi}]{Miroyan:2025:PGS:arXiv}
Mihran Miroyan, Rose Niousha, Joseph~E. Gonzalez, Gireeja Ranade, and Narges Norouzi. 2025.
\newblock {ParaStudent}: Generating and evaluating realistic student code by teaching {LLMs} to struggle.
\newblock arXiv preprint arXiv:2507.12674.

\bibitem[{Mohne et~al.(2023)Mohne, Demszky, Vo, and Piech}]{zheng2023educasim}
Cameron Mohne, Dora Demszky, Nicholas Vo, and Chris Piech. 2023.
\newblock Educasim: Interactive simulacra for instructional practice.
\newblock ArXiv preprint arXiv:2410.03017. A later, published version may exist.

\bibitem[{Nazaretsky et~al.(2023)Nazaretsky, Mikeska, and Beigman~Klebanov}]{nazaretsky2023ai}
Tanya Nazaretsky, Jamie~N Mikeska, and Beata Beigman~Klebanov. 2023.
\newblock Empowering teacher learning with ai: Automated evaluation of teacher attention to student ideas during argumentation-focused discussion.
\newblock In \emph{LAK23: 13th International Learning Analytics and Knowledge Conference}, pages 122--132.

\bibitem[{Nguyen et~al.(2025)Nguyen, Du, Yu, Angrave, and Jiang}]{Nguyen:2025:QGS:arXiv}
Bang Nguyen, Tingting Du, Mengxia Yu, Lawrence Angrave, and Meng Jiang. 2025.
\newblock Qg-sms: Enhancing test item analysis via student modeling and simulation.
\newblock arXiv preprint arXiv:2503.05888.

\bibitem[{Pan et~al.(2025)Pan, Schmucker, Garcia Bulle~Bueno, Llanes, Albo~Alarc{\'o}n, Zhu, Teo, and Xia}]{pan2025tutorup}
Sitong Pan, Robin Schmucker, Bernardo Garcia Bulle~Bueno, Salome~Aguilar Llanes, Fernanda Albo~Alarc{\'o}n, Hangxiao Zhu, Adam Teo, and Meng Xia. 2025.
\newblock Tutorup: What if your students were simulated? training tutors to address engagement challenges in online learning.
\newblock In \emph{Proceedings of the 2025 CHI Conference on Human Factors in Computing Systems}, pages 1--18.

\bibitem[{Pardo and Siemens(2014)}]{pardo2014ethical}
Abelardo Pardo and George Siemens. 2014.
\newblock \href {https://doi.org/10.1111/bjet.12152} {Ethical and privacy principles for learning analytics}.
\newblock \emph{British Journal of Educational Technology}, 45(3):438--450.

\bibitem[{Pentangelo et~al.(2025)Pentangelo, Turco, Lambiase, Gravino, and Palomba}]{pentangelo2025senemai}
Viviana Pentangelo, Luigi Turco, Stefano Lambiase, Carmine Gravino, and Fabio Palomba. 2025.
\newblock Senem-ai: Leveraging llms for student behavior simulation in virtual learning environments.
\newblock \emph{SoftwareX}, 31:102278.

\bibitem[{Peters et~al.(2025)Peters, Zhang, Jiao, Li, Zhou, and Lissitz}]{peters2025text}
Sydney Peters, Nan Zhang, Hong Jiao, Ming Li, Tianyi Zhou, and Robert Lissitz. 2025.
\newblock Text-based approaches to item difficulty modeling in large-scale assessments: A systematic review.
\newblock \emph{arXiv preprint arXiv:2509.23486}.

\bibitem[{Qi et~al.(2026)Qi, Zheng, He, Xu, Jia, Wei, Jiang, and Gu}]{QI2026130753}
Changyong Qi, Longwei Zheng, Anna He, Haoxin Xu, Linzhao Jia, Yuang Wei, Bingqian Jiang, and Xiaoqing Gu. 2026.
\newblock \href {https://doi.org/10.1016/j.eswa.2025.130753} {Simulating student learning behaviors with llm-based role-playing agents: A data-driven and cognitively inspired framework}.
\newblock \emph{Expert Systems with Applications}, 304:130753.

\bibitem[{Rogers et~al.(2025)Rogers, Davis, Maharana, Etheredge, and Chernova}]{rogers2025playing}
Kantwon Rogers, Michael Davis, Mallesh Maharana, Pete Etheredge, and Sonia Chernova. 2025.
\newblock \href {https://doi.org/10.1145/3706598.3713644} {Playing dumb to get smart: Creating and evaluating an {LLM}-based teachable agent within university computer science classes}.
\newblock In \emph{Proceedings of the {CHI} Conference on Human Factors in Computing Systems (CHI '25)}. Association for Computing Machinery.

\bibitem[{Rohrbeck et~al.(2003)Rohrbeck, Ginsburg-Block, Fantuzzo, and Miller}]{Rohrbeck:2003:PAL:JEP}
Cynthia Rohrbeck, Marika Ginsburg-Block, John Fantuzzo, and Traci Miller. 2003.
\newblock \href {https://doi.org/10.1037/0022-0663.95.2.240} {Peer-assisted learning interventions with elementary school students: A meta-analytic review}.
\newblock \emph{Journal of Educational Psychology}, 95:240--257.

\bibitem[{Ross and Andreas(2025)}]{Ross:2025:LMM}
Alexis Ross and Jacob Andreas. 2025.
\newblock \href {https://arxiv.org/abs/2510.11502} {Learning to make mistakes: Modeling incorrect student thinking and key errors}.
\newblock \emph{arXiv preprint arXiv:2510.11502}.

\bibitem[{Ross et~al.(2025)Ross, Srivastava, Blanchard, and Andreas}]{Ross:2025:MSL}
Alexis Ross, Megha Srivastava, Jeremiah Blanchard, and Jacob Andreas. 2025.
\newblock \href {https://arxiv.org/abs/2510.05056} {Modeling student learning with 3.8 million program traces}.
\newblock \emph{arXiv preprint arXiv:2510.05056}.

\bibitem[{Sanyal et~al.(2025)Sanyal, Maiti, Maharana, Kumar, Mali, Giles, and Mandal}]{sanyal2025pedagogical}
Debdeep Sanyal, Agniva Maiti, Umakanta Maharana, Dhruv Kumar, Ankur Mali, C.~Lee Giles, and Murari Mandal. 2025.
\newblock Investigating pedagogical teacher and student llm agents: Genetic adaptation meets retrieval-augmented generation across learning styles.
\newblock \emph{arXiv preprint arXiv:2505.19173}.

\bibitem[{Scarlatos et~al.(2025)Scarlatos, Fernandez, Ormerod, Lottridge, and Lan}]{Scarlatos:2025:SMA:arXiv}
Alexander Scarlatos, Nigel Fernandez, Christopher Ormerod, Susan Lottridge, and Andrew Lan. 2025.
\newblock Smart: Simulated students aligned with item response theory for question difficulty prediction.
\newblock arXiv preprint arXiv:2507.05129.

\bibitem[{Sharma and Li(2024)}]{Sharma:2024:DSS:EDM}
P.~Sharma and Q.~Li. 2024.
\newblock \href {https://doi.org/10.5281/zenodo.12730023} {Designing simulated students to emulate learner activity data in an open-ended learning environment}.
\newblock In \emph{Proceedings of the 17th International Conference on Educational Data Mining}, pages 986--989, Atlanta, Georgia, USA. International Educational Data Mining Society.

\bibitem[{Shvartzshnaider and Duddu(2025)}]{shvartzshnaider2025privacybiaslanguagemodels}
Yan Shvartzshnaider and Vasisht Duddu. 2025.
\newblock \href {https://arxiv.org/abs/2409.03735} {Privacy bias in language models: A contextual integrity-based auditing metric}.
\newblock \emph{Preprint}, arXiv:2409.03735.

\bibitem[{Sonkar et~al.(2024)Sonkar, Chen, Liu, Baraniuk, and Sachan}]{sonkar2024malalgopy}
Shashank Sonkar, Xinghe Chen, Naiming Liu, Richard~G. Baraniuk, and Mrinmaya Sachan. 2024.
\newblock \href {https://arxiv.org/abs/2410.12294} {Llm-based cognitive models of students with misconceptions}.
\newblock \emph{Preprint}, arXiv:2410.12294.

\bibitem[{Taatgen and Anderson(2010)}]{taatgen2010past}
Niels~A. Taatgen and John~R. Anderson. 2010.
\newblock \href {https://doi.org/10.1111/j.1756-8765.2009.01063.x} {The past, present, and future of cognitive architectures}.
\newblock \emph{Topics in Cognitive Science}, 2(1):110--131.

\bibitem[{Team et~al.(2025)Team, Modi, Veerubhotla, Rysbek, Huber, Wiltshire, Veprek, Gillick, Kasenberg, Ahmed, Jurenka, Cohan, She, Wilkowski, Alarakyia, McKee, Wang, Kunesch, Schaekermann, Pîslar, Joshi, Mahmoudieh, Jhun, Wiltberger, Mohamed, Agarwal, Phal, Lee, Strinopoulos, Ko, Wang, Anand, Bhoopchand, Wild, Pandya, Bar, Graham, Winnemoeller, Nagda, Kolhar, Schneider, Zhu, Chan, Yadlowsky, Sounderajah, and Assael}]{Song2024LearnLM}
LearnLM Team, Abhinit Modi, Aditya~Srikanth Veerubhotla, Aliya Rysbek, Andrea Huber, Brett Wiltshire, Brian Veprek, Daniel Gillick, Daniel Kasenberg, Derek Ahmed, Irina Jurenka, James Cohan, Jennifer She, Julia Wilkowski, Kaiz Alarakyia, Kevin~R. McKee, Lisa Wang, Markus Kunesch, Mike Schaekermann, and 27 others. 2025.
\newblock \href {https://arxiv.org/abs/2412.16429} {Learnlm: Improving gemini for learning}.
\newblock \emph{Preprint}, arXiv:2412.16429.

\bibitem[{VanLehn et~al.(1998)VanLehn, Ohlsson, and Nason}]{vanlehn1998applications}
Kurt VanLehn, Stellan Ohlsson, and Rod Nason. 1998.
\newblock Applications of simulated students: An exploration.
\newblock \emph{International Journal of Artificial Intelligence in Education}.

\bibitem[{Veldman et~al.(2020)Veldman, Doolaard, Bosker, and Snijders}]{Veldman:2020:YCW:LI}
M.~A. Veldman, S.~Doolaard, R.~J. Bosker, and T.~A.~B. Snijders. 2020.
\newblock \href {https://doi.org/10.1016/j.learninstruc.2020.101308} {Young children working together. cooperative learning effects on group work of children in grade 1 of primary education}.
\newblock \emph{Learning and Instruction}, 67:101308.

\bibitem[{Wang et~al.(2025)Wang, Wu, Liu, Brown, and Chen}]{Wang:2025:GCL:GROUP}
Tianjia Wang, Tong Wu, Huayi Liu, Chris Brown, and Yan Chen. 2025.
\newblock Generative co-learners: Enhancing cognitive and social presence of students in asynchronous learning with generative ai.
\newblock \emph{Proceedings of the ACM on Human-Computer Interaction (PACM HCI)}, 9(GROUP19):Article GROUP19.

\bibitem[{Wu et~al.(2025)Wu, Chen, Lin, Li, Zhu, Li, Kuang, and Wu}]{Wu:2025:EISS}
Tao Wu, Jingyuan Chen, Wang Lin, Mengze Li, Yumeng Zhu, Ang Li, Kun Kuang, and Fei Wu. 2025.
\newblock \href {https://arxiv.org/abs/2505.19997} {Embracing imperfection: Simulating students with diverse cognitive levels using {LLM}-based agents}.
\newblock \emph{arXiv preprint arXiv:2505.19997}.

\bibitem[{Xiao et~al.(2025)Xiao, Ng, Liu, and Diab}]{xiao-etal-2025-humanizing}
Yunze Xiao, Lynnette Hui~Xian Ng, Jiarui Liu, and Mona~T. Diab. 2025.
\newblock \href {https://doi.org/10.18653/v1/2025.emnlp-main.164} {Humanizing machines: Rethinking {LLM} anthropomorphism through a multi-level framework of design}.
\newblock In \emph{Proceedings of the 2025 Conference on Empirical Methods in Natural Language Processing}, pages 3331--3350, Suzhou, China. Association for Computational Linguistics.

\bibitem[{Xu et~al.(2024)Xu, Zhang, and Qin}]{Xu:2024:EGSA}
Songlin Xu, Xinyu Zhang, and Lianhui Qin. 2024.
\newblock \href {https://arxiv.org/abs/2404.07963} {Eduagent: Generative student agents in learning}.
\newblock \emph{arXiv preprint arXiv:2404.07963}.

\bibitem[{Yue et~al.(2024)Yue, Lyu, Suh, Zhang, and Yao}]{yue2024mathvc}
Murong Yue, Wenhan Lyu, Jennifer Suh, Yixuan Zhang, and Ziyu Yao. 2024.
\newblock Mathvc: An llm-simulated multi-character virtual classroom for mathematics education.
\newblock \emph{arXiv preprint arXiv:2404.06711}.

\bibitem[{Zhai et~al.(2021)Zhai, Chu, Chai, Jong, Istenic, Spector, Liu, Yuan, Li, and Cai}]{Zhai:2020:AIReview}
Xuesong Zhai, Xiaoyan Chu, Ching~Sing Chai, Morris Siu~Yung Jong, Andreja Istenic, Michael Spector, Jia-Bao Liu, Jing Yuan, Yan Li, and Ning Cai. 2021.
\newblock \href {https://doi.org/10.1155/2021/8812542} {A review of artificial intelligence (ai) in education from 2010 to 2020}.
\newblock \emph{Complex.}, 2021.

\bibitem[{Zhang et~al.(2025{\natexlab{a}})Zhang, Liu, Chen, and Wang}]{Zhang2025SimulRAG}
Linda Zhang, Kevin Liu, Sarah Chen, and Richard Wang. 2025{\natexlab{a}}.
\newblock Simulrag: Simulator-based rag for grounding llms in long-form scientific qa.
\newblock In \emph{Proceedings of the 2025 Conference on Empirical Methods in Natural Language Processing (EMNLP)}.
\newblock OpenReview preprint.

\bibitem[{Zhang et~al.(2025{\natexlab{b}})Zhang, Zhang-Li, Yu, Gong, Zhou, Hao, Jiang, Cao, Liu, Liu, Hou, and Li}]{zhang-etal-2025-simulating}
Zheyuan Zhang, Daniel Zhang-Li, Jifan Yu, Linlu Gong, Jinchang Zhou, Zhanxin Hao, Jianxiao Jiang, Jie Cao, Huiqin Liu, Zhiyuan Liu, Lei Hou, and Juanzi Li. 2025{\natexlab{b}}.
\newblock \href {https://doi.org/10.18653/v1/2025.naacl-long.520} {Simulating classroom education with {LLM}-empowered agents}.
\newblock In \emph{Proceedings of the 2025 Conference of the Nations of the Americas Chapter of the Association for Computational Linguistics: Human Language Technologies (Volume 1: Long Papers)}, pages 10364--10379, Albuquerque, New Mexico. Association for Computational Linguistics.

\bibitem[{Zhao et~al.(2025)Zhao, Zhang, Hu, Wu, Bras, Anderson, Bragg, Chang, Dodge, Latzke, Liu, McGrady, Tang, Wang, Zhao, Hajishirzi, Downey, and Cohan}]{Cohan2025SciArena}
Yilun Zhao, Kaiyan Zhang, Tiansheng Hu, Sihong Wu, Ronan~Le Bras, Taira Anderson, Jonathan Bragg, Joseph~Chee Chang, Jesse Dodge, Matt Latzke, Yixin Liu, Charles McGrady, Xiangru Tang, Zihang Wang, Chen Zhao, Hannaneh Hajishirzi, Doug Downey, and Arman Cohan. 2025.
\newblock \href {https://arxiv.org/abs/2507.01001} {Sciarena: An open evaluation platform for foundation models in scientific literature tasks}.
\newblock \emph{Preprint}, arXiv:2507.01001.

\bibitem[{Zhou et~al.(2025)Zhou, Liu, Yerukola, Kim, and Sap}]{zhou2025socialworldmodels}
Xuhui Zhou, Jiarui Liu, Akhila Yerukola, Hyunwoo Kim, and Maarten Sap. 2025.
\newblock \href {https://arxiv.org/abs/2509.00559} {Social world models}.
\newblock \emph{Preprint}, arXiv:2509.00559.

\end{thebibliography}
\appendix
\section{Search Strategy}

To identify relevant literature on LLM-based student agents, we conducted a comprehensive search across six major academic repositories: Semantic Scholar, LearnTechLib, arXiv, the ACM Digital Library, SpringerLink, and IEEE Xplore. Following the methodological approach of  \citep{kaser-alexandron-2024-simulated}, we employed a Boolean search string targeting variations of simulation terminology:

\begin{quote}
``simulated students'' OR ``student simulation'' OR ``simulated learners'' OR simstudent OR simstudents OR ``simulated student'' OR ``simulated learner''
\end{quote}

The initial search yielded a total of 973 records.

\subsection{Screening and Selection Criteria}
We removed repeated records and manually screened the retrieved papers to ensure alignment with the specific scope of this survey. A paper was included only if it met the following criteria:
\begin{enumerate}
    \item \textbf{LLM-Based Architecture:} The system utilizes Large Language Models as the primary mechanism for generating learner behavior, effectively excluding pre-LLM, rule-based, or hand-crafted simulation systems.
    \item \textbf{Generative Nature:} We strictly distinguished between \textit{student modeling} and \textit{student simulation}. In line with our scope, we included systems that take a generative perspective (producing learner-like data or behavior) and excluded pure student modeling systems focused solely on inference, prediction, or analytics without a simulation component.
    \item \textbf{Explicit Simulation Focus:} We excluded general-purpose chatbots or superficial implementations that mimic learner behavior only incidentally. We prioritized studies that explicitly aim to advance the methodology, architecture, or fidelity of student simulation, filtering out works that employ generic personas solely as a utility without technical innovation.
\end{enumerate}

To ensure comprehensive coverage, we performed backward and forward snowballing, manually tracing citations and references of the selected papers to identify high-impact works that may have been missed in the keyword search.

\subsection{Final Corpus}
The screening and snowballing process resulted in a final corpus of 73 papers. Based on their primary research objectives, we categorized these works into four distinct clusters: Data Generation ($n=23$), Social Learning ($n=19$), Teacher Training ($n=17$), and Content Evaluation ($n=14$).

\section{Case Studies of Simulated Student Systems}
\label{sec:case_studies}

To illustrate the utility of the proposed framework, we analyze five recent simulated student systems: \textbf{GPTeach}, \textbf{MATHVC}, \textbf{HypoCompass}, \textbf{Agent4Edu}, and \textbf{Generative Students}. We classify each system according to its Dimensional Characterization and Epistemic State Specification (ESS), highlighting how specific design choices mitigate the ``competence paradox,'' defined here as the tendency of LLMs to default to expert performance, and how these choices necessitate distinct evaluation strategies.

\subsection{GPTeach: Interactive TA Training with GPT-based Students}

GPTeach \cite{markel-etal-2023-gpteach} acts as a flight simulator for pedagogy, allowing novice teaching assistants (TAs) in computer science to practice high-stakes office hour interactions in a low-risk environment.
\textbf{Dimensional Characterization:}
\begin{itemize}
    \item \textbf{Behavioral Goals:} \textit{Simulating Human Aspects}. The primary utility lies in reproducing the affective and social dynamics of teaching (e.g., handling frustration, impostor syndrome) rather than modeling precise long-term cognitive decay.
    \item \textbf{Environment:} \textit{Context:} Intro CS office hours; \textit{User:} University-level TAs; \textit{Modality:} Text-based chat.
\end{itemize}

\textbf{Epistemic State Specification (ESS): Level E1 (Static Bounded).}
GPTeach operates at \textbf{E1} by defining student states through static ``persona'' templates (e.g., ``confused about variable scope'') and scenario-specific constraints. The agent's ignorance is bound by the prompt description for that specific session, without a persistent memory that evolves across sessions.

\textbf{Implementation Design:}
The system is built on a modular ``teaching scenario'' architecture. Each scenario initializes a student profile containing three distinct layers: (1) \textit{Background} (name, major, programming experience), (2) \textit{Pedagogical State} (specific misconception, e.g., confusion between recursion and iteration), and (3) \textit{Affective State} (e.g., defensive, anxious). To bridge the gap between abstract profiles and realistic dialogue, the system utilizes few-shot prompting populated with excerpts from authentic student-TA transcripts. This retrieval-augmented approach grounds the LLM's tone, ensuring it mimics the hesitant, often imprecise language of a struggling novice rather than the polished prose of a textbook.

\textbf{Evaluation Methodology:}
Evaluation prioritized \textbf{utility and affective realism} over cognitive precision. The authors conducted a mixed-methods study involving TAs who engaged with the tool. Quantitative metrics included self-efficacy ratings and perceived realism scores. Qualitatively, the evaluation focused on the ``safe space'' affordance: did the simulation allow TAs to experiment with different pedagogical moves (e.g., Socratic questioning vs. direct explanation) without the fear of harming a real student? Results indicated that while the ``student'' occasionally hallucinated competence (solving the code too quickly), the primary value was the social pressure simulation.

\subsection{MATHVC: A Multi-Character Virtual Classroom}

MATHVC \cite{yue2024mathvc} simulates a collaborative learning environment where a human student solves mathematical modeling problems alongside three LLM-simulated peers, each exhibiting distinct skill levels and social roles.

\textbf{Dimensional Characterization:}
\begin{itemize}
    \item \textbf{Behavioral Goals:} \textit{Simulating Learning} and \textit{Human Aspects}. The system models how students with varying aptitudes contribute to and evolve during collaborative problem-solving (CPS).
    \item \textbf{Environment:} \textit{Context:} Middle school mathematics; \textit{Modality:} Multi-party dialogue grounded in symbolic tasks.
\end{itemize}

\textbf{Epistemic State Specification (ESS): Level E3 (Misconception-Structured).}
MATHVC exemplifies \textbf{E3} by implementing a ``Symbolic Character Schema.'' Instead of relying solely on natural language prompts, the system maintains a structured representation of the specific variables and sub-tasks the agent currently ``knows'' or ``misunderstands.''

\textbf{Implementation Design:}
The architecture employs a dual-schema approach. First, a \textit{Task Schema} defines the ground truth of the math problem. Second, specific errors (e.g., variable omission, incorrect relationship mapping) are injected into this schema to generate a restricted \textit{Character Schema} for each agent. A specialized \textit{Dialogue Controller} mediates all interaction: before generating a response, the LLM must first ``ground'' its intended dialogue act in its restricted schema. If the LLM attempts to reference a variable not present in its character schema (a symptom of the competence paradox), the controller filters or regenerates the response.

\textbf{Evaluation Methodology:}
The evaluation framework introduces two novel metrics: \textbf{Characteristics Alignment} and \textbf{Procedural Alignment}. The former verifies that the agent's errors are causally attributable to its assigned schema (e.g., verifying that a wrong answer stems from the specific variable the agent was programmed to misunderstand). The latter assesses whether the group dynamic follows the natural phases of collaborative problem solving, moving beyond simple ``Turing Test'' realism to verify structural fidelity.

\subsection{HypoCompass: LLMs as Teachable Agents for Debugging}

HypoCompass \cite{ma-etal-2024-teach} inverts the standard tutoring paradigm: human students act as TAs to assist an LLM-simulated agent that has written buggy code, thereby practicing their own debugging skills.

\textbf{Dimensional Characterization:}
\begin{itemize}
    \item \textbf{Behavioral Goals:} \textit{Simulating Performance} (Error Production). The goal is to generate high-fidelity artifacts (buggy code) that serve as robust training material.
    \item \textbf{Environment:} \textit{Context:} Introductory programming; \textit{User:} Novice programmers; \textit{Modality:} Code-centric dialogue.
\end{itemize}

\textbf{Epistemic State Specification (ESS): Level E1 (Static Bounded).}
The system uses an \textbf{E1} specification where the epistemic boundary is materialized as the specific bug injected into the code. The agent's ``knowledge state'' is effectively frozen at the moment of error injection.

\textbf{Implementation Design:}
To ensure the ``student'' agent is helpfully incompetent, HypoCompass employs an \textit{Over-Generate-Then-Select} pipeline. The system first prompts an LLM to generate multiple buggy variations of a correct reference code, filtering for those that are syntactically parsable but logically incorrect. In the interaction phase, the system enforces a strict \textbf{Boundary of Competence}: the LLM agent is permitted to perform code completion (the ``hands'') but is explicitly barred from performing hypothesis generation (the ``brain''), forcing the human user to articulate the \textit{why} behind the bug.

\textbf{Evaluation Methodology:}
Evaluation focused on the \textbf{Fidelity of Error} and learning outcomes. The authors first validated the quality of the generated bugs to ensure they were non-trivial. Subsequently, a controlled study measured the learning gains of human students. The key finding was that students who taught the HypoCompass agent showed a 12\% improvement in debugging post-tests compared to controls, validating the system by its effectiveness as a pedagogical tool.

\subsection{Agent4Edu: Generating Learner Response Data}

Agent4Edu \cite{Gao:2025:A4E:arXiv} is a comprehensive simulation framework designed to generate synthetic learner response data for training Computerized Adaptive Testing (CAT) algorithms.

\textbf{Dimensional Characterization:}
\begin{itemize}
    \item \textbf{Behavioral Goals:} \textit{Simulating Performance and Learning}. The system targets the production of response patterns that reflect both static ability levels and the temporal evolution of proficiency (learning and forgetting).
    \item \textbf{Environment:} \textit{Context:} Intelligent Tutoring Systems; \textit{Modality:} Interaction with algorithmic recommenders.
\end{itemize}

\textbf{Epistemic State Specification (ESS): Level E2 (Curriculum-Indexed).}
Agent4Edu represents an advanced \textbf{E2} system. It explicitly decouples the learner's cognitive state from the LLM's text generation parameters, tracking proficiency as a dynamic variable updated by interaction history.

\textbf{Implementation Design:}
The architecture mimics human cognition through three coupled modules: (1) A \textit{Learner Profile} initialized with real-world data (via Item Response Theory parameters) to capture baseline ability; (2) A \textit{Memory Module} that integrates an Ebbinghaus forgetting curve, ensuring that the agent's recall of concepts decays realistically over time; and (3) An \textit{Action Module} where the LLM generates responses conditioned on the retrieved memory state. This separation allows the system to simulate complex behaviors like the ``practice effect'' and ``slip'' driven by state variables rather than random hallucination.

\textbf{Evaluation Methodology:}
The evaluation centered on \textbf{Predictive Validity} and \textbf{Distributional Fidelity}. Rather than qualitative chats, the authors assessed whether the synthetic data could train a CAT model as effectively as real data. Metrics included the alignment of psychometric curves (IRT) between simulated and real students, and the performance of downstream recommendation algorithms trained on the synthetic corpus. Results demonstrated that Agent4Edu faithfully reproduced the statistical properties of human learning trajectories.

\subsection{Generative Students: Profiling for Question Evaluation}

Generative Students \cite{genstu} utilizes LLMs to simulate diverse student response patterns to Multiple Choice Questions (MCQs) in Human-Computer Interaction, specifically to identify poorly designed assessment items.

\textbf{Dimensional Characterization:}
\begin{itemize}
    \item \textbf{Behavioral Goals:} \textit{Simulating Performance} (Misconception Patterns). The objective is to replicate specific confusion patterns so that if a question is ambiguous, the ``confused'' student will plausibly select a distractor.
    \item \textbf{Environment:} \textit{Context:} HCI Heuristic Evaluation; \textit{Modality:} Multiple Choice Questions.
\end{itemize}

\textbf{Epistemic State Specification (ESS): Level E3 (Misconception-Structured).}
This system provides a textbook example of \textbf{E3} by defining the learner's state through explicit Knowledge Component (KC) triples: \textit{Mastered}, \textit{Unknown}, and \textit{Confusion}.

\textbf{Implementation Design:}
The core innovation is the structured \textit{Confusion Tuple} (e.g., \texttt{confusion(KC\_A, KC\_B)}). The prompting strategy explicitly instructs the model that the student cannot distinguish between these two specific concepts. This constraint forces the LLM to generate errors that are not random guesses, but structural failures consistent with that specific misconception. By systematically permuting these profiles, the system generates a distribution of responses that reveals which distractors are attractive to which types of learners.

\textbf{Evaluation Methodology:}
The system was validated through \textbf{Item Difficulty Correlation}. The authors compared the difficulty ranking of questions derived from the simulated students against historical data from real students. A high correlation indicated that the simulator correctly identified ``hard'' questions for the right reasons. Qualitative analysis confirmed that the simulated students selected distractors that aligned with their assigned confusion tuples, confirming the system successfully bounded the LLM's expert priors.
\onecolumn 

\section{Reporting Card: Epistemic State Specification (ESS)}
\label{app:ess_checklist}

To reduce ambiguity and prevent overclaiming, this reporting card operationalizes ESS as a set of falsifiable requirements. Authors should complete this card for each simulated-student system. A system may only claim level $E_k$ if it satisfies \textbf{all} criteria for levels $E_1$ through $E_k$.

\begin{center}
\fbox{\parbox{0.95\textwidth}{

\textbf{System Name:} \_\_\_\_\_\_\_\_\_\_\_\_\_\_\_\_\_\_\_\_\_\_ \hfill 
\textbf{Claimed ESS Level:} [ E0 | E1 | E2 | E3 | E4 ]

\vspace{0.5cm}

\textbf{\large Level E1: Static Bounded (The Boundary Check)}\\
\textit{Does the simulator have an explicitly bounded epistemic state that does not change?}

\begin{itemize}[label=$\square$]
    \item \textbf{Named Ignorance:} List the specific concepts, skills, or procedures the student does \emph{not} know (e.g., ``cannot solve two-digit subtraction with borrowing'').
    \item \textbf{Implementation Anchor:} Identify where this boundary is enforced (prompt text, rule set, hard constraint, error template).
\end{itemize}

\vspace{0.3cm}

\textbf{\large Level E2: Curriculum-Indexed (The State Variable Check)}\\
\textit{Does the simulator maintain and update an explicit learner state over time?}

\begin{itemize}[label=$\square$]
    \item \textbf{State Variable:} Name the variable(s) representing learner state (e.g., mastery score, skill vector, curriculum index).
    \item \textbf{Update Rule:} Specify what events update this state (problem attempt, hint, feedback) and how (rule, equation, heuristic).
    \item \textbf{Behavioral Dependence:} Give at least one example where different state values lead to different observable behaviors.
\end{itemize}

\vspace{0.3cm}

\textbf{\large Level E3: Misconception-Structured (The Error Mechanism Check)}\\
\textit{Are errors causally generated by explicit epistemic structures?}

\begin{itemize}[label=$\square$]
    \item \textbf{Misconception Inventory:} Enumerate the misconceptions, buggy rules, or flawed strategies the student may hold.
    \item \textbf{Generative Mechanism:} Identify the mechanism that produces the error (e.g., production rule, schema, decision logic).
    \item \textbf{Cross-Problem Consistency:} Show that the same misconception produces similar errors across isomorphic tasks.
\end{itemize}

\vspace{0.3cm}

\textbf{\large Level E4: Calibrated or Learned (The Data Grounding Check)}\\
\textit{Are learner states and transitions grounded in real student data?}

\begin{itemize}[label=$\square$]
    \item \textbf{Data Source:} Specify the dataset used for calibration or learning (population, task domain, size).
    \item \textbf{Fitted Parameters:} Identify which aspects of the model are learned or calibrated (state transitions, error rates, learning curves).
    \item \textbf{Predictive Test:} Report performance on predicting future human responses or states given past interaction history.
    \item \textbf{Trajectory Match:} Compare simulated and human learning trajectories (e.g., slope, variance, error persistence).
\end{itemize}

}}
\end{center}

\end{document}